\documentclass[letterpaper]{article}
\usepackage{aaai19}
\usepackage{times}
\usepackage{helvet}
\usepackage{courier}
\usepackage{url}
\usepackage{xcolor}
\usepackage{graphicx}
\usepackage{amsmath}
\usepackage{quoting}
\usepackage{xcolor}
\newcommand\qdata{\textsc{QuaRel}}
\newcommand\qsystem{\textsc{QuaSP}}
\newcommand\qsystemplus{\textsc{QuaSP+}}
\newcommand\frictiondata{\textsc{QuaRel$^F$}}
\newcommand\qsystemdenotation{\textsc{BiLSTM}}
\newcommand\qsystemzero{\textsc{QuaSP+Zero}}

\newcommand{\namecite}[1]{\citeauthor{#1}~\shortcite{#1}}

\newcommand{\eat}[1]{}
\newenvironment{ite}
  {\parskip 0cm \begin{itemize} \parsep 0cm \itemsep 0cm \topsep 0cm}
  {\end{itemize}}
\newenvironment{enu}
  {\parskip 0cm \begin{list}{}{\parsep 0cm \itemsep 0cm \topsep 0cm}}
  {\end{list}}

\newenvironment{myquote}
  {\parskip 1mm \begin{quoting}[vskip=1mm,leftmargin=5mm]}
  {\end{quoting}}

\title{\qdata: A Dataset and Models for \\ 
Answering Questions about Qualitative Relationships}
\author{Oyvind Tafjord, Peter Clark, Matt Gardner, Wen-tau Yih, Ashish Sabharwal \\
Allen Institute for AI, Seattle, WA \\
\{oyvindt,peterc,mattg,scottyih,ashishs\}@allenai.org 
}

\begin{document}

\maketitle

\begin{abstract}
Many natural language questions require recognizing and reasoning with qualitative
relationships (e.g., in science, economics, and medicine), but are challenging to
answer with corpus-based methods. Qualitative modeling provides tools that
support such reasoning, but the semantic parsing task of
mapping questions into those models has formidable challenges.
We present \qdata{}, a dataset of diverse story questions involving
qualitative relationships that characterize these challenges, and techniques that begin to address them.
The dataset has 2771 questions relating 19 different types of quantities. For example, \emph{``Jenny observes that the robot vacuum cleaner
moves slower on the living room carpet than on the bedroom carpet. Which carpet has more friction?''}
We contribute (1) a simple and flexible conceptual framework for representing these kinds of questions;
(2) the \qdata~dataset, including logical forms, exemplifying the parsing challenges; and
(3) two novel models for this task, built as extensions of type-constrained semantic parsing. The
first of these models (called \qsystemplus) significantly outperforms off-the-shelf tools on~\qdata.
The second (\qsystemzero) demonstrates zero-shot capability, i.e., the ability to handle
new qualitative relationships without requiring additional training data, something
not possible with previous models. This work thus makes inroads into answering complex,
qualitative questions that require reasoning, and scaling to new relationships at low cost. The dataset and models are available at http://data.allenai.org/quarel.
\end{abstract}

\section{Introduction}

\begin{figure}
\centerline{
\fbox{
  \parbox{\columnwidth}{
    {\bf Qualitative Story Problem:} \\    
       Alan noticed that his toy car \textcolor{teal}{rolls further} on a \textcolor{red}{wood floor} than on a \textcolor{blue}{thick carpet}. This suggests that:\\
\hspace*{5mm}   (A) \textcolor{blue}{The carpet} has \textcolor{orange}{more resistance} \\
\hspace*{5mm}    (B) \textcolor{red}{The floor} has \textcolor{orange}{more resistance} \\
        {\bf Solution:} (A) The carpet has more resistance \\
        {\bf Identification of worlds being compared:} \\
        \hspace*{5mm}{\includegraphics[width=0.75\columnwidth]{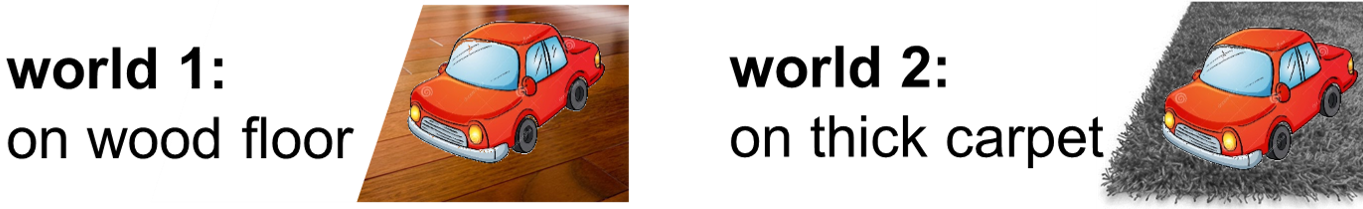}} \\
          {\bf Question Interpretation (Logical Form):} \\
    qrel(\textcolor{teal}{distance, higher}, \textcolor{red}{world1}) $\rightarrow$ \\
    \hspace*{5mm} qrel(\textcolor{orange}{friction, higher}, \textcolor{blue}{world2}) ; \\
    \hspace*{5mm} qrel(\textcolor{orange}{friction, higher}, \textcolor{red}{world1})?
    }
}}
\caption{An example problem from \qdata~and its logical form (LF), from which the
  answer can be inferred (Section~\ref{representation}).
  The problem is conceptualized as comparing two {\it worlds}
  which the semantic parser needs to identify and track. 
  In the LF, qrel($p$, higher$\vert$lower, $w$) denotes that $p$ is higher/lower in world $w$ (compared with the other world). Colors show approximate correspondence between the question and the LF.}
\label{example}
\end{figure}

Many natural language tasks require recognizing and reasoning with qualitative relationships.
For example, we may read about temperatures rising (climate science), a drug dose
being increased (medicine), or the supply of goods being reduced (economics), and want to
reason about the effects. Qualitative story problems, of the kind found in elementary
exams (e.g., Figure~\ref{example}), form a natural example of many of these 
linguistic and reasoning challenges, and is the target of this work.

Understanding and answering such questions is particularly challenging.
Corpus-based methods perform poorly in this setting, as the questions ask about novel scenarios
rather than facts that can be looked up. Similarly, word association methods struggle, as a single word
change (e.g., ``more'' to ``less'') can flip the answer. Rather, the task
appears to require {\it knowledge of the underlying qualitative relations}
(e.g., ``more friction implies less speed'').

Qualitative modeling \cite{Forbus1984QualitativePT,weld2013readings,kuipers1994qualitative} provides a means
for encoding and reasoning about such relationships. Relationships are expressed
in a natural, qualitative way (e.g., if X increases, then so will Y), rather than requiring numeric equations,
and inference allows complex questions to be answered.
However, the semantic parsing task of mapping real world questions
into these models is formidable and presents unique challenges.
These challenges must be solved if natural questions
involving qualitative relationships are to be reliably answered.

We make three contributions: (1) a simple and flexible conceptual framework for
formally representing these kinds of questions,
in particular ones that express qualitative comparisons between two scenarios; (2) a challenging new dataset (\qdata{}), including
logical forms, exemplifying the parsing challenges; and
(3) two novel models that extend type-constrained semantic parsing to address these challenges.  

Our first model, \qsystemplus, addresses the problem of tracking different ``worlds'' in questions,
resulting in significantly higher scores than with off-the-shelf tools (Section~\ref{world-tracking}).
The second model, \qsystemzero, demonstrates zero-shot capability, i.e., the ability to handle
new qualitative relationships on unseen properties, without requiring additional training data,
something not possible with previous models (Section~\ref{zero-shot}).
Together these contributions make inroads into answering complex, qualitative questions by
linking language and reasoning, and offer a new dataset and models to spur further progress by the community.

\section{Related Work}
\label{related-work}
There has been rapid progress in question-answering (QA), spanning a wide
variety of tasks and phenomena, including factoid QA \cite{rajpurkar2016squad},
entailment \cite{snli}, sentiment \cite{maas-EtAl:2011:ACL-HLT2011}, and ellipsis and coreference \cite{Long2016SimplerCL}.
Our contribution here is the first dataset specifically targeted at qualitative
relationships, an important category of language that has been less explored.
While questions requiring reasoning about qualitative relations 
sometimes appear in other datasets, e.g., \cite{Clark2018ThinkYH}, our dataset
specifically focuses on them so their challenges can be studied.

For answering such questions, we treat the problem as mapping language to a
structured formalism (semantic parsing) where simple qualitative
reasoning can occur. Semantic parsing has a long history~\cite{zelle1996learning,Zettlemoyer2005LearningTM,berant2013semantic,Krishnamurthy2017NeuralSP}, using datasets about geography~\cite{zelle1996learning}, travel booking~\cite{dahl1994expanding}, factoid QA over knowledge bases~\cite{berant2013semantic}, Wikipedia tables~\cite{pasupat2015compositional}, and many more.
Our contributions to this line of research are: a dataset that
features phenomena under-represented in prior datasets, namely (1) highly diverse
language describing open-domain qualitative problems, and (2) the need to
reason over entities that have no explicit formal representation; and methods
for adapting existing semantic parsers to address these phenomena.

For the target formalism itself, we draw on the extensive body of work on
qualitative reasoning \cite{Forbus1984QualitativePT,weld2013readings,kuipers1994qualitative} to create a logical form language
that can express the required qualitative knowledge,
yet is sufficiently constrained that parsing into it
is feasible, described in more detail in Section~\ref{representation}.

There has been some work connecting language with qualitative reasoning,
although mainly focused on extracting qualitative models themselves from text
rather than question interpretation, e.g., \cite{mcfate2014using,mcfate2016scaling}.
Recent work by \namecite{crouse2018learning} also includes interpreting
questions that require identifying qualitative processes in text,
in constrast to our setting of interpreting NL story questions
that involve qualitative comparisons.

Answering story problems has received attention in the
domain of arithmetic, where simple algebra story questions (e.g., ``Sue had
5 cookies, then gave 2 to Joe...'') are mapped to a system of equations,
e.g., \cite{Ling2017ProgramIB,Kushman2014LearningTA,Wang2017DeepNS,shi2015automatically}.
This task is loosely analogous to ours (we instead map to qualitative relations)
except that in arithmetic the entities to relate are often
identifiable (namely, the numbers). Our qualitative story
questions lack this structure, adding an extra challenge.

The \qdata{} dataset shares some structure with the Winograd Schema Challenge \cite{levesque2011winograd}, being 2-way multiple choice questions invoking both commonsense and coreference. However, they test different aspects of commonsense: Winograd uses coreference resolution to test commonsense understanding of scenarios, while \qdata{} tests reasoning about qualitative relationships requiring tracking of coreferent ``worlds.''

Finally, crowdsourcing datasets has become a driving force in AI,
producing significant progress, e.g., \cite{rajpurkar2016squad,JoshiTriviaQA2017,Wang2015BuildingAS}.
However, for semantic parsing tasks, one obstacle has been the
difficulty in crowdsourcing target logical forms for questions.
Here, we show how those logical forms can be obtained indirectly
from workers without training the workers in the formalism, loosely
similar to \cite{Yih2016TheVO}.

\section{Knowledge Representation \label{representation}}

We first describe our framework for representing questions and the knowledge
to answer them. Our dataset, described later,
includes logical forms expressed in this language.

\subsection{Qualitative Background Knowledge}

We use a simple representation of qualitative relationships, leveraging prior work in
qualitative reasoning \cite{Forbus1984QualitativePT}.
Let $P = \{p_{i}\}$ be the set of properties relevant to the question set's domain (e.g., smoothness, friction, speed).
Let $V_{i} = \{v_{ij}\}$ be a set of qualitative values for property $p_{i}$ (e.g., fast, slow).
For the {\it background knowledge} about the domain itself (a qualitative model), following \namecite{Forbus1984QualitativePT},
we use the following predicates:
\begin{myquote}
  q+({\it property1}, {\it property2}) \\
  q-({\it property1}, {\it property2})
\end{myquote}
q+ denotes that {\it property1} and {\it property2} are qualitatively proportional,
e.g., if {\it property1} goes up, {\it property2} will too, while q- denotes inverse proportionality, e.g.,
\begin{myquote}
{\it \# If friction goes up, speed goes down.} \\
q-(friction, speed).
\end{myquote}
We also introduce the predicate:
\begin{myquote}
  higher-than($val_{ij}$, $val_{ik}$, {\it property}$_{i}$)
\end{myquote}
where $val_{ij} \in V_{i}$, allowing an ordering of property values to be specified, e.g., higher-than(fast, slow, speed).
For our purposes here, we simplify to use just two property values, low and high, for
all properties. (The parser learns mappings from words to these values, described later).

\begin{figure}
    \centering
    \includegraphics[width=\columnwidth]{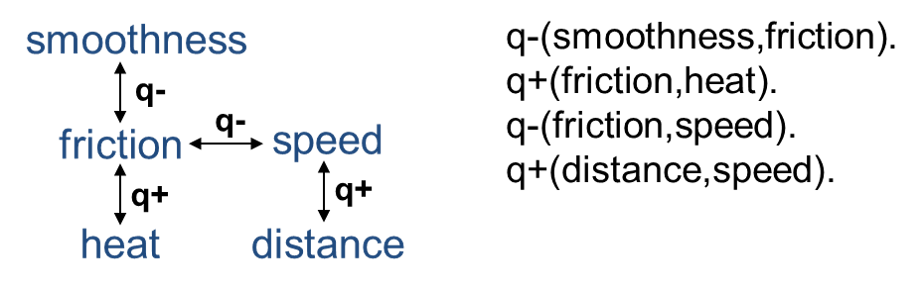}     
    \caption{A simple qualitative theory about friction, shown graphically (left) and formally (right). For example, \mbox{q-(smoothness,friction)} indicates that if smoothness increases, friction decreases.
    \label{friction}}
\end{figure}

Given these primitives, compact theories can be authored for a particular domain
by choosing relevant properties $P$, and specifying qualitative relationships (q+,q-)
and ordinal values (higher-than) for them.
For example, a simple theory about friction is sketched graphically in Figure~\ref{friction}.
Our observation is that these theories are relatively small,
simple, and easy to author.
Rather, the primary challenge is in mapping the complex and varied
language of questions into a form that interfaces with this representation.

This language can be extended
to include additional primitives from qualitative modeling,
e.g., i+(x,y) (``the {\it rate of change} of x is qualitatively
proportional to y''). That is, 
the techniques we present are not specific to our particular
qualitative modeling subset. The only requirement is that, given a set
of absolute values or qualitative relationships from a question,
the theory can compute an answer.

\begin{figure*}[t]
\centerline{
\fbox{%
  \parbox{\textwidth}{%
    \begin{enu}
    \setlength{\itemsep}{3pt}
\item[(1)] Heather wants to see if a bar stool will slide faster along the bar surface which has decorative raised bumps on it or on the smooth wooden floor. On which surface will the chair slide faster? (A) bar (B) floor\\
{\small LF: \textit{qval(smoothness, low, world1), qval(smoothness, high, world2) $\rightarrow$ qrel(speed, higher, world1) ; qrel(speed, higher, world2)}}
\item[(2)] Andy was running across the tile floor and sliding across it. There was less friction here, but he thought he could do the same outside on the cement. When Andy tries to slide across the cement, his socks will make \textunderscore \textunderscore \textunderscore \textunderscore \textunderscore{} than when he slides across the tile floor. (A) more heat (B) less heat\\
{\small LF: \textit{qrel(friction, lower, world1) $\rightarrow$ qrel(heat, higher, world2 ; qrel(heat, lower, world2)}}
\item[(3)] Mary noticed that erasing her mistakes on her drawing paper seemed to take more effort than the marker paper.  This caused more heat to develop on (A) the drawing paper or (B) the marker paper\\
{\small LF: \textit{qrel(friction, higher, world1) $\rightarrow$ qrel(heat, higher, world1) ; qrel(heat, higher, world2)}}
\item[(4)] Henry is playing with his younger brother.  Henry is bigger and stronger and he can throw the ball (A) farther (B) not as far.
{\small LF: \textit{qrel(strength, higher, world1) $\rightarrow$ qrel(distance, higher, world1) ; qrel(distance, lower, world1)}}
\item[(5)] It's turkey hunting season and Jim is on his front porch. He hears a gun shot off the the west. Then he hears another one off to the north. The one to the north was easier to hear than the one to the west. Which hunter is closer to Jim's house? (A) the one to the west (B) the one to the north\\
{\small LF: \textit{qrel(loudness, higher, world1) $\rightarrow$ qrel(distance, lower, world2) ; qrel(distance, lower, world1)}}
    \end{enu}
  }
}}
\caption{Examples of questions and logical forms in the \qdata~dataset (the first 3 are also in the friction subset, \frictiondata{})}
\label{turked-examples}
\end{figure*}

\subsection{Representing Questions}

\subsubsection{Predicates.}
A key feature of our representation is the conceptualization of questions as
describing events happening in two {\it worlds}, world1 and world2, that are being compared.
That comparison may be between two different entities, or the same entity at different
time points. E.g., in Figure~\ref{example} the two worlds being compared are the car on wood,
and the car on carpet. The tags world1 and world2 denote these different situations,
and semantic parsing (Section~\ref{systems}) requires learning to correctly associate
these tags with parts of the question describing those situations. This abstracts
away irrelevant details of the worlds, while still
keeping track of which world is which.

We define the following two predicates to express qualitative information in questions:
\begin{myquote}
  qrel({\it property, direction, world}) \\
  qval({\it property, value, world})
\end{myquote}
where
  {\it property} ($p_i$) $\in$ P, 
  {\it value} $\in$ $V_{i}$,
  {\it direction} $\in$ \{higher, lower\}, and 
  {\it world} $\in$ \{world1, world2\}.
    {\bf qrel()} denotes the relative assertion that {\it property} is {\it higher/lower} in {\it world}
    compared with the other world, which is left implicit,\footnote{
  We consider just two worlds being compared here, but the formalism generalizes
  to N-way comparisons by adding a fourth argument: qrel({\it prop, dir, world, other-world}).}
  e.g., from Figure~\ref{example}: 
\begin{myquote}
  {\it \# The car rolls further on wood.} \\
  qrel(distance, higher, world1)
\end{myquote}
where world1 is a tag for the ``car on wood'' situation (hence world2 becomes
a tag for the opposite ``car on carpet'' situation).
{\bf qval()} denotes that {\it property} has an absolute {\it value} in {\it world}, e.g.,
\begin{myquote}
  {\it \# The car's speed is slow on carpet.} \\
  qval(speed, low, world2)
\end{myquote}

\subsection{Logical Forms for Questions}

\label{sec:logical-forms}

Despite the wide variation in language, the space of logical forms (LFs) for the
questions that we consider is relatively compact. In each question, the question body establishes
a scenario and each answer option then probes an implication. 
We thus express a question's LF as a tuple:
\begin{myquote}
  {\it (setup, answer-A, answer-B)}
\end{myquote}
where {\it setup} is the predicate(s) describing the scenario, and {\it answer-*} are the
predicate(s) being queried for. If {\it answer-A} follows from {\it setup}, as inferred by
the reasoner, then the answer is (A); similarly for (B). For readability we will write this as
\begin{myquote}
  {\it setup} $\rightarrow$ {\it answer-A} ; {\it answer-B}
\end{myquote}
We consider two styles of LF, covering a large range of questions. The
first is:
\begin{myquote}
  (1) qrel($p,d,w$) $\rightarrow$ \\
  \hspace*{8mm} qrel($p',d',w'$) ; qrel($p',d'',w''$)
\end{myquote}
which deals with relative values of properties between worlds, and applies when the question setup includes a comparative.
An example of this is in Figure~\ref{example}. The second is:
\begin{myquote}
  (2) qval($p,v,w$), qval($p,v',w'''$) $\rightarrow$ \\
  \hspace*{8mm} qrel($p',d',w'$) ; qrel($p',d'',w''$) 
\end{myquote}
which deals with absolute values of properties, and applies when the setup uses absolute terms instead of comparatives.
An example is the first question in Figure~\ref{turked-examples}, shown simplified below, whose LF looks as follows (colors showing approximate correspondences):
\begin{myquote}
{\it \# Does a bar stool \textcolor{orange}{slide faster} along the \textcolor{red}{bar} surface with \textcolor{teal}{decorative raised bumps} or the \textcolor{magenta}{smooth wooden} \textcolor{blue}{floor}? (A) \textcolor{red}{bar} (B) \textcolor{blue}{floor}} \vspace{1mm} \\ 
    qval(\textcolor{teal}{smoothness, low}, \textcolor{red}{world1}), \\
    qval(\textcolor{magenta}{smoothness, high}, \textcolor{blue}{world2}) $\rightarrow$ \\
    \hspace*{5mm} qrel(\textcolor{orange}{speed, higher}, \textcolor{red}{world1}) ; \\
    \hspace*{5mm} qrel(\textcolor{orange}{speed, higher}, \textcolor{blue}{world2})
\end{myquote}

\subsection{Inference}

A small set of rules for qualitative reasoning connects these predicates together.
For example, (in logic) {\bf if} the value of P is higher in world1 than the value of P in world2 {\bf and} q+(P,Q) {\bf then} the value of Q will be higher in world1 than the value of Q in world2.
Given a question's logical form, a qualitative model, and these rules, a Prolog-style inference engine determines which answer option follows from the premise.\footnote{
  E.g., in Figure~\ref{example}, the qualitative model includes q-(friction, distance), and
  the general qualitative reasoning rules include opposite(world1, world2) and
  qrel(P, D, W) $\wedge$ q-(P, P') $\wedge$ opposite(W, W') $\rightarrow$ qrel(P', D, W'),
  so the answer can be inferred.}

\section{The \qdata~Dataset \label{dataset}}
\label{dataset-description}

\qdata{} is a crowdsourced dataset of 2771 multiple-choice story questions, including
their logical forms. The size of the dataset is similar to several other datasets with annotated logical
forms used for semantic parsing~\cite{zelle1996learning,Hemphill1990TheAS,Yih2016TheVO}.
As the space of LFs is constrained, the dataset
is sufficient for a rich exploration of this space.

We crowdsourced multiple-choice questions in two parts, encouraging workers
to be imaginative and varied in their use of language.
First, workers were given a seed qualitative
relation q+/-($p_{1},p_{2}$) in the domain, expressed in English
(e.g., ``If a surface has more friction, then an object will travel slower''),
and asked to enter two objects, people, or situations to compare.
They then created a question, guided by a large number of examples,
and were encouraged to be imaginative and use their own words. The results are a remarkable variety of
situations and phrasings (Figure~\ref{turked-examples}).

Second, the LFs were elicited using a novel technique of reverse-engineering them
from a set of follow-up questions, without exposing workers to the underlying
formalism. This is possible because of the constrained space of LFs.
Referring to LF templates (1) and (2) earlier (Section~\ref{sec:logical-forms}), these questions are as follows:
\begin{enumerate}
\item[1.] What is the correct answer (A or B)?
\item[2.] Which property are the answer options asking about? ($p'\in~\{p_{1},p_{2}\}$)
\item[3.] In the correct answer, is this property higher or lower than in the incorrect answer? ($d'$)
\item[4.] Do the answer options:
  \begin{ite}
  \item ask the {\it same} question about {\it different} objects/situations? ($d'=d'',w'\neq~w''$)
  \item ask {\it opposite} questions about {\it the same} object/situation? ($d'\neq~d'',w'=w''$)
  \end{ite}
\item[5.] Which direction of comparison is used in the body of the question?
  \begin{ite}
  \item higher/lower? ($d$, LF template is (1))
  \item {\it OR} were two values given? If so, enter the values, standardized as high/low in the LF ($v$,$v'$, LF template is (2))
  \end{ite}
\end{enumerate}
From this information, we can deduce the target LF ($p$ is the complement
of $p'\in~\{p_{1},p_{2}\}$, $w'''\neq~w$, we arbitrarily set $w$=world1, hence all other
variables can be inferred). Three independent workers answer these follow-up
questions to ensure reliable results.

We also had a human answer the questions in the dev partition (in principle, they should all be 
answerable). The human scored 96.4\%, the few failures caused
by occasional annotation errors or ambiguities in the question set itself,
suggesting high fidelity of the content.

About half of the dataset are questions about friction, relating five different properties 
(friction, heat, distance, speed, smoothness). These questions form a meaningful, connected subset of the dataset
which we denote \frictiondata{}. The remaining questions involve a wide variety of 14 additional properties
and their relations, such as ``exercise intensity vs.\ sweat'' or ``distance vs.\ brightness''.\footnote{See supplementary material at http://data.allenai.org/quarel for a complete list.}

\begin{table}[t]
  \centering
  \begin{tabular}{|l|c|} \hline
    {\bf Property} & {\bf Size} \\ \hline
    \# questions & 2771 \\
    \# questions (\frictiondata{}~subset) & 1395 \\
    \# type (1), (2) (see Section~\ref{sec:logical-forms}) & 2048, 723 \\
    \# questions train/dev/test & 1941/278/552 \\
    Min/avg/max qn length (words) & 11/37/112 \\
    Min/avg/max qn length (sent.) & 1/2.4/8 \\
    Vocab (\# uniq words) & 5226 \\ \hline
  \end{tabular}
  \caption{Summary statistics for the \qdata~dataset. \label{statistics}}
\end{table}

\begin{figure}
\centerline{\small
\fbox{%
  \parbox{0.47\textwidth}{%
``smoother'' (864),
``rougher'' (568),
``more smooth'' (270),
``more rough'' (126),
``bumpier'' (55),
``less bumpy'' (27),
``more rugged'' (24),
``easier'' (16),
``not as rough'' (16),
``flatter'' (10),
``slicker'' (10),
``isn't as rough'' (8),
``stickier'' (8),
``rugged'' (7),
``more even'' (6),
``more easily'' (5),
``spikier'' (4),
``more uniform'' (4),
``softer'' (4),
``more level'' (4),
``bumpier ride'' (4),
``glide more easily'' (4),
``more jagged'' (4),
``lumpier'' (3),
``more fluidly'' (3),
``rolls easily'' (2),
``struggle to move'' (2),
``ideal surface'' (2),
``more freely'' (2),
``harder'' (2),
``has more ridges'' (2),
``more uniform surface'' (2),
``more scraping'' (2),
``more ridges'' (2),
``sleek'' (2),
``easier to pull'' (1),
``barely a touch'' (1),
``flatter surface'' (1),
``much better'' (1),
``full of bumps'' (1),
``rocky'' (1),
``calmer'' (1),
``less slick'' (1),
``bumpiness'' (1),
``less obstacles'' (1),
``more silky glide'' (1),
``sanded'' (1),
``slick'' (1),
``lots of bumps'' (1),
``not nearly as smooth'' (1),
``easier to wipe'' (1),
``snagging'' (1),
``easier on my feet'' (1),
``rolls more easily'' (1)
    }
}}
\caption{Examples of the varied way that smoother/rougher surfaces are described in \qdata~questions.}
\label{phrase-examples}
\end{figure}

Figure~\ref{turked-examples} shows typical examples of questions in \qdata,
and Table~\ref{statistics} provides summary statistics. In particular, the
vocabulary is highly varied (5226 unique words), given the dataset size.
Figure~\ref{phrase-examples} shows some examples of the varied phrases
used to describe smoothness.

\section{Baseline Systems \label{systems}}

We use four systems to evaluate the difficulty of this dataset. 
(We subsequently present two new models, extending the baseline neural semantic
parser, in Sections~\ref{world-tracking} and~\ref{zero-shot}). The first two
are an information retrieval system and a word-association method, following
the designs of \namecite{Clark2016CombiningRS}. These are naive baselines that
do not parse the question, but nevertheless may find some signal in a large corpus of text
that helps guess the correct answer. The third is a CCG-style rule-based semantic parser written
specifically for friction questions (the \frictiondata~subset), but prior to data being collected.  
The last is a state-of-the-art neural semantic parser.  We briefly describe each in turn.

\subsubsection{Information Retrieval (IR) System}
  To answer multiple-choice questions, this system searches a large (280GB) text corpus
  to see if the question $q$ along with an answer option is loosely stated in the corpus,
  and returns the confidence that such a statement was found. To do this, for each answer option $a_i$,
  it sends $q + a_i$ as a query to a search engine and returns the search engine's score for 
  the top retrieved sentence $s$ where $s$ also has at least one non-stopword overlap with $q$,
  and at least one with $a_i$. The option with the highest score is selected.
  
\subsubsection{Pointwise Mutual Information (PMI)}

  Word co-occurrences may also provide some signal for answering these questions,
  e.g., the high co-occurrence of ``faster'' and ``ice'' in a corpus may help answer a
  question ending with ``...faster? (A) ice (B) gravel''. To formalize this,
  given a question $q$ and an answer option $a_i$, we use PMI \cite{church1989} to measure
  the strength of the associations between parts of $q$ and parts of $a_i$. Given a large corpus $C$,
  the PMI for two n-grams $x$ and $y$ is defined as
\begin{equation*}
\mathrm{PMI}(x,y) = \log \frac{p(x,y)}{p(x) p(y)}
\end{equation*}
The system selects the answer with the largest average PMI, calculated over all pairs of question n-grams and answer option n-grams.

\subsubsection{Rule-based Semantic Parser}

The rule-based semantic parser uses a simple CCG-like grammar \cite{steedman2011combinatory} specifically written for the
friction scenario task (\frictiondata{})
over several days, but prior to the dataset being constructed. It represents a good-faith attempt to
solve this subset of questions with traditional methods. First, the question is preprocessed to
tag likely references to the worlds being compared, using hand-written rules that look for
surface names (``road'', ``ice''), appropriate adjectives (``rough'', ``green''), and
by position (``over $<$X$>$''). The first candidate word/phrase is tagged world1
(with type WORLD), the second world2, and if those phrases occur later in the question,
they are tagged with the corresponding world. The system then parses the question using
142 task-specific, CCG-like rules, such as: \vspace{1mm}

\noindent
\hspace*{2mm}  ``is greater than'' $\vdash$ (S$\backslash$PROPERTY)$\backslash$WORLD:\\
\hspace*{3cm}  $\lambda$p.$\lambda$w.qrel(p, HIGHER, w) \\
\hspace*{12mm} ``velocity'' $\vdash$ PROPERTY:speed   \vspace{1mm} 

\noindent
where $\backslash$WORLD means ``look left for something of category WORLD''. Thus a tagged phrase
like
\begin{myquote}
  ``the velocity on ice[world2] is greater than'' 
\end{myquote}
produces qrel(speed, higher, world2). The parser skips over most words in the
story, only paying attention to words that are tagged or referenced in the grammar. 
\begin{figure}
\begin{center}
{\includegraphics[width=\columnwidth]{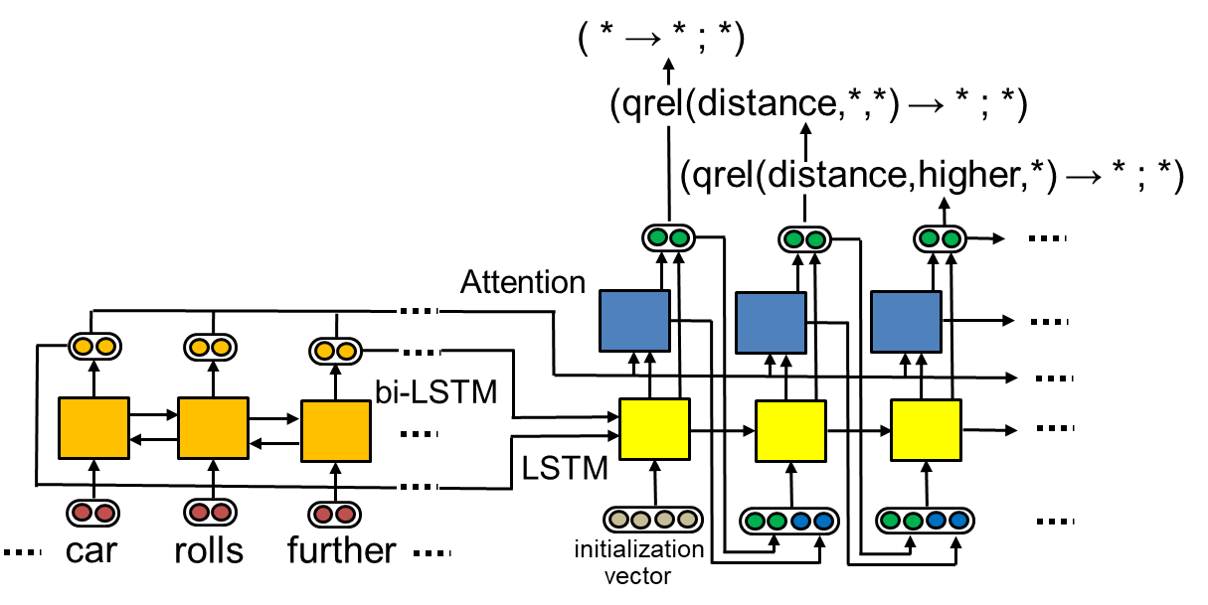}}
\end{center}
\caption{The \qsystem~parser decodes to a sequence of LF-building decisions, incrementally constructing the LF by selecting production rules from the LF grammar. As illustrated, first it decides if the LF should be of type 1 or 2 (here, type 1 is chosen), then it selects the the property for the question body (here, distance), then it selects the direction of change (here, higher), and so on.}
\label{model}
\end{figure}

\subsubsection{Type-constrained Neural Semantic Parser (\qsystem)} 
Our final system is AllenNLP's implementation of a neural semantic parser~\cite{gardner2018allennlp}.
This parser uses a type-constrained, encoder-decoder architecture, representative of the current
state-of-the-art on many datasets~\cite{Krishnamurthy2017NeuralSP,Yin2017ASN,goldman2017weakly}.
The model architecture is similar to standard seq2seq models, with an LSTM that
encodes the question and an LSTM with attention over the encoded question that decodes a logical form.
However, unlike standard seq2seq models that output logical form tokens directly, this parser
outputs production rules from a CFG-like grammar over the space of all logical forms.
These production rules sequentially build up an abstract syntax tree, which determines the
logical form.  In this way, the parser is constrained to only produce valid
LFs, and does not have to spend modeling capacity learning the syntax of
the language.

For our domain, we created a simple grammar capturing the logical form language described in Section~\ref{sec:logical-forms}.
The parser uses this grammar to find the set of valid choices at each step of decoding.
The model architecture, with example inputs and outputs, is illustrated in Figure~\ref{model}.  We refer to this instantiation of
the parser as \qsystem.  As \qdata{} has annotated logical forms, this model is trained to maximize the likelihood of the logical
form associated with each question. At test time, beam search is used to find the highest scoring parse.

As input to the model we feed the full question plus answer options as a single sequence of tokens, encoding each token using a concatenation of Glove \cite{pennington2014glove} and ELMo \cite{Peters2018DeepCW} vectors.

As a separate baseline, we also train a similar two-layer 
bi-directional LSTM encoder (\qsystemdenotation{} in the results)
to directly predict answer A vs.\ B, without an intermediate logical form.\footnote{For more implementation details, see supplementary material at http://data.allenai.org/quarel.}

\section{Baseline Experiments}
\label{baselines}
We ran the above systems on the \qdata{} dataset.
\qsystem{} was trained on the training set, using the model with highest parse accuracy on the dev set 
(similarly \qsystemdenotation{} used highest answer accuracy on the dev set) .
The results are shown in Table~\ref{results}. 
The 95\% confidence interval is +/- 4\% on the full test set. The human score
is the sanity check on the dev set (Section~\ref{dataset}).

\begin{table}
  \centering
  \begin{tabular}{|l|ll|ll|} \hline
    {\bf Dataset $\rightarrow$} & \multicolumn{2}{|c|}{\bf \qdata} & \multicolumn{2}{|c|}{\bf \frictiondata} \\
    {\bf Model $\downarrow$} & {\bf Dev} & {\bf Test}  & {\bf Dev} & {\bf Test}\\ \hline
    Random & 50.0 & 50.0 & 50.0 & 50.0 \\
    Human & 96.4 & - & 95.0 & - \\
    IR & 50.7 & 48.6 & 50.7 & 48.9 \\
    PMI & 49.3 & 50.5 & 50.7 & 52.5 \\
    Rule-Based & - & - & 55.0 & 57.7 \\
    \qsystemdenotation & 55.8 & 53.1 & 59.3 & 54.3 \\
    \qsystem & 62.1 & 56.1 & 69.2 & 61.7 \\ \hline
    \qsystemplus & {\bf 68.9} & {\bf 68.7} & {\bf 79.6} & {\bf 74.5} \\ \hline
  \end{tabular}
  \caption{Scores (answer accuracies) of the different models on the full \qdata~dataset
    and \frictiondata~subset about friction.
    The baseline models only marginally outperform a random baseline.
    In \qsystemplus, however, identifying and delexicalizing the worlds
    significantly improves the performance (see Section~\ref{world-tracking}).}
  \label{results}
\end{table}

As Table~\ref{results} shows, the \qsystem{}
model performs better than other baseline approaches
which are only slightly above
random. \qsystem{} scores 56.1\% (61.7\% on the friction subset), indicating the challenges of this dataset.

For the rule-based system, we observe that it is unable to parse the majority (66\%)
of questions (hence scoring 0.5 for those questions, reflecting a random guess), due to the 
varied and unexpected vocabulary present in the dataset. For example, Figure~\ref{phrase-examples} shows some of the
ways that the notion of ``smoother/rougher'' is expressed in
questions, many of which are not covered by the hand-written CCG grammar.
This reflects the typical brittleness of hand-built systems.

\begin{table}
  \centering
  \begin{tabular}{|l|cc|cc|} \hline
    {\bf Dataset $\rightarrow$} & \multicolumn{2}{|c|}{\bf \qdata} & \multicolumn{2}{|c|}{\bf \frictiondata} \\
    {\bf Model $\downarrow$} & {\bf Dev} & {\bf Test}  & {\bf Dev} & {\bf Test}\\ \hline
\qsystem  & 37.4 & 32.2 & 47.9 & 43.2 \\ \hline
\qsystemplus & {\bf 46.8} & {\bf 43.8} & {\bf 64.3} & {\bf 59.0} \\ \hline
  \end{tabular}
  \caption{Parse accuracies for the semantic parsers.}
    \label{parse-accuracy}
\end{table}

For \qsystem, we also analyzed the parse accuracies, shown in Table~\ref{parse-accuracy},
the score reflecting the percentage of times it produced exactly the right
logical form. The random baseline for parse accuracy is near zero given the large space of logical forms, while the model parse accuracies are relatively high, much better than a random baseline.

Further analysis of the predicted LFs indicates that the neural model
does well at predicting the properties ($\sim$25\% of errors on dev set), but struggles 
to predict the worlds in the LFs reliably ($\sim$70\% of errors on dev set). 
This helps explain why non-trivial parse accuracy
does not necessarily translate into 
correspondingly higher answer accuracy: If only the world assignment is wrong, the answer will flip and give a score of zero, rather than the average 0.5.

\section{New Models \label{new-models}}

We now present two new models, both extensions of the neural baseline~\qsystem.
The first,~\qsystemplus, addresses the leading cause of failure just described, namely
the problem of identifying the two worlds being compared, and significantly outperforms
all the baseline systems. The second,~\qsystemzero, addresses the
scaling problem, namely the costly requirement of needing many training examples
each time a new qualitative property is introduced. It does this by instead
using only a small amount of lexical information about the new property,
thus achieving ``zero shot'' performance, i.e., handling properties unseen in the training examples  \cite{Palatucci2009ZeroshotLW},
a capability not present in the baseline systems. We present the models and results for each.

\subsection{\qsystemplus: A Model Incorporating World Tracking}
\label{world-tracking}

We define the world tracking problem as identifying and tracking
references to different ``worlds'' being compared in text,
i.e., correctly mapping phrases to world identifiers, a 
critical aspect of the semantic parsing task.
There are three reasons why this is challenging. First, 
unlike properties, the worlds being compared in questions
are {\it distinct in almost every question}, and thus there
is no obvious, learnable mapping from phrases to worlds.
For example, while a property (like speed)
has learnable ways to refer to it (``faster'', ``moves rapidly'',
``speeds'', ``barely moves''), worlds are different
in each question (e.g., ``on a road'', ``countertop'',
``while cutting grass'') and thus learning to identify them is hard.
Second, different phrases may be used to refer to the
same world in the same question (see Figure~\ref{different-world-expressions}),
further complicating the task.
Finally, even if the model could learn to identify worlds in other ways, e.g., by syntactic
position in the question, there is the problem of 
selecting world1 or world2 consistently throughout the
parse, so that the equivalent phrasings are assigned the same world.

This problem of mapping phrases to world identifiers 
is similar to the task of entity linking \cite{Ling2015DesignCF}.
In prior semantic parsing work, entity linking is relatively straightforward: simple string-matching
heuristics are often sufficient~\cite{jia2016data,dong2016language}, or
an external entity linking system can be used~\cite{Yih2015SemanticPV,xu2016question}.
In \qdata, however, because the phrases denoting world1 and world2 are different in almost every question, and the word ``world'' is never used, such methods cannot
be applied.

To address this, we have developed ~\qsystemplus, a new model
that extends~\qsystem{} by adding an extra initial step 
to identify and delexicalize world references in the question.
In this delexicalization process, potentially new linguistic descriptions of worlds are replaced by {\it canonical tokens},
creating the opportunity for the model to generalize across questions.
For example, the world mentions in the question:
\begin{myquote}
  ``A ball rolls further on wood than carpet because the (A) carpet is smoother (B) wood is smoother''
  \end{myquote}
are delexicalized to:
\begin{myquote}
  ``A ball rolls further on \textsc{World1} than \textsc{World2} because the (A) \textsc{World2} is smoother (B) \textsc{World1} is smoother''  
\end{myquote}
This approach is analogous to \namecite{Herzig2018DecouplingSA}, who delexicalized words to POS tags to avoid memorization.
Similar delexicalized features have also been employed in Open Information Extraction \cite{Etzioni2008OpenIECACM},
so the Open IE system could learn a {\it general} model of how relations are expressed.
In our case, however, delexicalizing to \textsc{World1} and \textsc{World2} is itself a significant challenge,
because identifying phrases referring to worlds is substantially more complex than (say) identifying parts of speech.

To perform this delexicalization step, we use the world annotations included as part of the training dataset (Section~\ref{dataset-description})
to train a separate tagger to 
identify ``world mentions'' (text spans) in the question using BIO tags\footnote{
e.g., the world mention ``calm water'' in the question ``...in calm water, but...'' would be tagged ``...in/O calm/B water/I, but/O...''
} (BiLSTM encoder followed by a CRF).
The spans are then sorted into \textsc{World1} and \textsc{World2} using the following algorithm:
\begin{enumerate}
\item[1.] If one span is a substring of another, they are are grouped together. Remaining spans are singleton groups.
\item[2.] The two groups containing the longest spans are labeled as the two worlds being compared.
\item[3.] Any additional spans are assigned to one of these two groups based on closest edit distance (or ignored if zero overlap).
\item[4.] The group appearing first in the question is labeled \textsc{World1}, the other \textsc{World2}.
\end{enumerate}
The result is a question in which world mentions are canonicalized.
The semantic parser \qsystem~is then trained using these questions.\footnote{
  During training, using alignment with the annotations, we ensure the worlds in the LF are numbered consistently with these tags.
}
We call the combined system (delexicalization plus semantic parser) \qsystemplus. 

The results for \qsystemplus{} are included in Table~\ref{results}.
Most importantly, \qsystemplus{} significantly outperforms the baselines
by over 12\% absolute. Similarly, the parse accuracies are significantly
improved from 32.2\% to 43.8\% (Table~\ref{parse-accuracy}).
This suggests that this delexicalization technique is an effective way of making
progress on this dataset, and more generally on problems where
multiple situations are being compared, a common characteristic of
qualitative problems.

\begin{figure}
  \centering
  \small{
  \begin{tabular}{|c|} \hline
``road'' \& ``paved roadway'' \\
``wooden bar'' \& ``wood counter'' \\
``her counter is stone'' \& ``stone counter'' \\
``grass'' \& ``(mowing his) yard'' \\
``shag carpeting'' \& ``carpet'' \\
``tiled floor'' \& ``tile'' \\
``wet tennis court'' \& ``wet court'' \\
``wastebasket'' \& ``waste basket'' \\
``ice on the pond'' \& ``ice pond'' \\
``wood beam'' \& ``wooden beam'' \\
``outside'' \& ``grass'' \\
``street'' \& ``asphalt'' \\
``carpet'' \& ``carpeted floor'' \\
``hardwood'' \& ``wood'' \\
``beach'' \& ``sand'' \\
``mulch'' \& ``mulched area'' \\ \hline
  \end{tabular}
  }
  \caption{Examples of different linguistic expressions of the same world in a question.}
    \label{different-world-expressions}
\end{figure}

\subsection{\qsystemzero: A Model for the Zero-Shot Task}
\label{zero-shot}

While our delexicalization procedure demonstrates a way
of addressing the world tracking problem, 
the approach still relies on annotated data; if we were to
add new qualitative relations, new training data would be needed,
which is a significant scalability obstacle. To address this, 
we define the zero-shot problem as being able to answer questions involving a new predicate p given training data only about other predicates P different from p.
For example, if we add a new property (e.g., heat) to the qualitative model (e.g., adding q+(friction, heat); ``more friction implies more heat''), we want to answer questions involving heat without creating new annotated training questions, and instead only use minimal extra information about the new property.
A parser that achieved good zero-shot performance, i.e.,
worked well for new properties unseen at training time,
would be a substantial advance, allowing a new qualitative
model to link to questions with minimal effort. 

\qdata~provides an environment in which methods for this
zero-shot theory extension can be devised and evaluated.
To do this, we consider the following experimental setting:
All questions mentioning a particular property are removed,
the parser is trained on the remainder, and then tested
on those withheld questions, i.e., questions mentioning
a property unseen in the training data.

We present and evaluate a model that we have developed for this,
called \qsystemzero, that modifies the \qsystemplus~parser as follows:
During decoding, at points where the parser is selecting which
property to include in the LF (e.g., Figure~\ref{model}),
it does not just consider the question tokens, but also the
{\it relationship} between those tokens and the properties $P$
used in the qualitative model. For example, a question token
such as ``longer'' can act as a cue for (the property) length,
even if unseen in the training data, because ``longer'' and a lexical
form of length (e.g.,``length'') are similar.
This approach follows the entity-linking approach
used by \namecite{Krishnamurthy2017NeuralSP},
where the similarity between question tokens and
(words associated with) entities - called the entity linking score -
help decide which entities to include in the LF during parsing.
Here, we modify their entity linking score $s(p,i)$,
linking question tokens $q_i$ and property ``entities'' $p$, to be:
\begin{equation*}
s(p,i) = \max_{w \in W(p)} v^T_w K v_{q_i}
\end{equation*}
where $K$ is a diagonal matrix connecting the embedding of the
question token $q_i$ and words $W(p)$ associated with the property $p$.
For $W(p)$, we provide a small list of words for each property (such as ``speed'', ``velocity'',
and ``fast'' for the speed property), a small-cost requirement.

The results with \qsystemzero~are in Table~\ref{zero-shot-table}, shown in detail on the \frictiondata~subset and (due to space constraints) summarized for the full \qdata.
We can measure overall performance of \qsystemzero~by averaging each of the zero-shot test sets (weighted by the number of questions in each set), resulting in an overall parse accuracy of 38.9\% and answer accuracy 61.0\% on \frictiondata, and
25.7\% (parse) and 59.5\% (answer) on \qdata, both significantly better than random. 
These initial results are encouraging, suggesting that it may be possible to parse into modified qualitative models that include new relations, with minimal annotation effort, significantly opening up qualitative reasoning methods for QA.

\begin{table}
  \centering
 { \small
  \begin{tabular}{|l|cc|cc|} \hline
    {\bf Held out} & \multicolumn{2}{|c|}{\bf Parse (LF) Accuracy} 
    & \multicolumn{2}{|c|}{\bf Answer Accuracy} \\
    {\bf property} &  {\bf Seen} & {\bf Unseen} & {\bf Seen} & {\bf Unseen} \\ \hline
distance   & 46.7 & 33.3 & 67.0 & 59.8 \\ 
friction   & 54.4 & 29.7 & 78.6 & 59.4 \\
heat       & 33.5 & 52.6 & 58.8 & 64.7 \\
smoothness & 47.9 & 53.2 & 67.7 & 62.2 \\
speed      & 45.0 & 33.1 & 64.9 & 60.2 \\ \hline
None       & 51.8 &   NA & 66.7 &   NA \\ \hline
Weighted avg. & 44.5 & {\bf 38.9} & 66.2 & {\bf 61.0} \\ \hline
  \end{tabular}
  } 
 \caption{
   Baseline scores (bold) using \qsystemzero~for the zero-shot task of answering
   questions involving properties unseen in the training data, using
   the \frictiondata{}~subset of \qdata. For the entire
   \qdata~dataset, the weighted average scores for questions with unseen
   properties are {\bf 25.7\%} (parse) and {\bf 59.5\%} (answer).}
    \label{zero-shot-table}
\end{table}

\section{Summary and Conclusion}

Our goal is to answer questions that involve qualitative relationships, an
important genre of task that involves both language and knowledge,
but also one that presents significant challenges for semantic parsing.
To this end we have developed a simple and flexible formalism for
representing these questions; constructed \qdata, the first
dataset of qualitative story questions that exemplifies these challenges;
and presented two new models that adapt existing parsing techniques to this
task. The first model, \qsystemplus, illustrates how delexicalization can help
with world tracking (identifying different ``worlds'' in questions),
resulting in state-of-the-art performance on \qdata.
The second model, \qsystemzero, illustrates how zero-shot learning
can be achieved (i.e., adding new qualitative relationships without requiring new training examples)
by using an entity-linking approach applied to properties - a capability
not present in previous models. 

There are several directions in which this work can be expanded.
First, quantitative property values (e.g., ``10 mph'') are currently not handled well,
as their mapping to ``low'' or ``high'' is context-dependent. 
Second, some questions do not fit our two question templates (Section~\ref{sec:logical-forms}),
e.g., where two property values are a single answer option (e.g., ``....(A) one floor is smooth and the other floor is rough'').
Finally, some questions include an additional level of indirection, requiring an inference step to map to qualitative relations.
For example, ``Which surface would be best for a race? (A) gravel (B) blacktop'' requires the additional commonsense inference that ``best for a race'' implies ``higher speed''.

Given the ubiquity of qualitative comparisons in natural text,
recognizing and reasoning with qualitative relationships is likely
to remain an important task for AI. This work makes inroads into
this task, and contributes a dataset and models to encourage
progress by others. The dataset and models are publicly available at http://data.allenai.org/quarel.

\bibliography{references}
\bibliographystyle{aaai}
\end{document}